\documentclass[10pt,twocolumn,twoside]{IEEEtran}
\usepackage{cite}
\usepackage{graphicx}
\usepackage{epstopdf}
\ifCLASSINFOpdf
\else
\fi
\usepackage[cmex10]{amsmath}

\usepackage{algorithm}
\usepackage{algorithmic}

\usepackage{array}

\ifCLASSOPTIONcompsoc
 \usepackage[caption=false,font=normalsize,labelfont=sf,textfont=sf]{subfig}
\else
 \usepackage[caption=false,font=footnotesize]{subfig}
\fi
\usepackage{fixltx2e}

\usepackage{stfloats}
\usepackage{url}

\begin{document}
%
\title{Bayesian Compressive Sensing Using Normal Product Priors}

\author{Zhou Zhou, Kaihui Liu, and Jun Fang,\IEEEmembership{
Member,~IEEE}
\thanks{Zhou Zhou, Kaihui Liu, and Jun Fang are with the National Key Laboratory on Communications,
University of Electronic Science and Technology of China, Chengdu
611731, China, Email: ZhouZhou@std.uestc.edu.cn;~kaihuil@std.uestc.edu.cn;~JunFang@uestc.edu.cn}

\thanks{This work was supported in part by the National
Science Foundation of China under Grant 61172114.}}

\maketitle

\begin{abstract}
In this paper, we introduce a new sparsity-promoting prior,
namely, the ``normal product'' prior, and develop an efficient
algorithm for sparse signal recovery under the Bayesian framework.
The normal product distribution is the distribution of a product
of two normally distributed variables with zero means and possibly
different variances. Like other sparsity-encouraging distributions
such as the Student's $t$-distribution, the normal product
distribution has a sharp peak at origin, which makes it a suitable
prior to encourage sparse solutions. A two-stage normal
product-based hierarchical model is proposed. We resort to the
variational Bayesian (VB) method to perform the inference.
Simulations are conducted to illustrate the effectiveness of our
proposed algorithm as compared with other state-of-the-art
compressed sensing algorithms.
\end{abstract}


%
\IEEEpeerreviewmaketitle

\section{Introduction}
\IEEEPARstart{C}{ompressed} sensing
\cite{donoho2006compressed,candes2006robust} is a new data
acquisition technique that has attracted much attention over the
past decade. Existing methods for compressed sensing can generally
be classified into the following categories, i.e. the greedy
pursuit approach \cite{tropp2007signal}, the convex
relaxation-type approach \cite{chen1998atomic} and the nonconvex
optimization method \cite{chartrand2007exact}. Another class of
compressed sensing techniques that have received increasing
attention are Bayesian methods. In the Bayesian framework, the
signal is usually assigned a sparsity-encouraging prior, such as
the Laplace prior and the Gaussian-inverse Gamma prior
\cite{tipping2001sparse}, to encourage sparse solutions. It has
been shown in a series of experiments that Bayesian compressed
sensing techniques \cite{ji2008bayesian} demonstrate superiority
for sparse signal recovery as compared with the greedy methods and
the basis pursuit method. One of the most popular prior model for
Bayesian compressed sensing is the Gaussian-inverse Gamma prior
proposed in \cite{tipping2001sparse}. The Gaussian-inverse Gamma
prior is a two-layer hierarchical model in which the first layer
specifies a Gaussian prior to the sparse signal and an inverse
Gamma priori is assigned to the parameters characterizing the
Gaussian prior. As discussed in \cite{wipf2004sparse}, this
two-stage hierarchical model is equivalent to imposing a Student's
$t$-distribution on the sparse signal. Besides the
Gaussian-inverse Gamma prior, authors in
\cite{babacan2010bayesian} employ Laplace priors for the sparse
signal to promote sparse solutions.

In this paper, we introduce a new sparsity-encouraging prior,
namely, the normal product (NP) prior, for sparse signal recovery.
A two-stage normal product-based hierarchical model is
established, and we resort to the variational Bayesian (VB) method
to perform the inference for the hierarchical model. Our
experiments show that the proposed algorithm achieves similar
performance as the sparse Bayesian learning method, while with
less computational complexity.


\section{The Bayesian Network}\label{bayesian network}
\begin{figure}[!b]
\centering
\includegraphics[width=7.5cm,height=5cm]{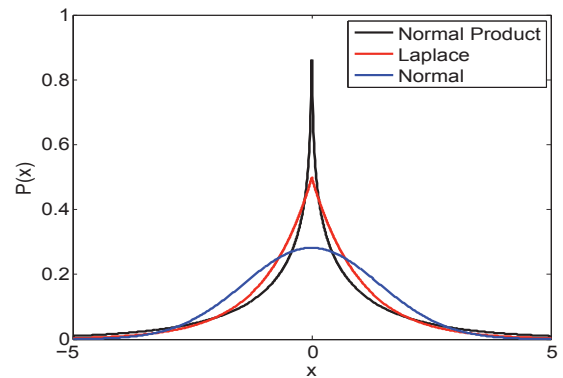}
\caption{Three kinds of PDF with the same standard deviation}
\label{bayesian network graph}
\end{figure}

In the context of compressed sensing, we are given a noise corrupted linear measurements of a vector ${\bf {x}}_0$
\begin{equation}\label{CS}
{\bf{y}} = {\bf{A}}{{\bf{x}}_{{0}}} + {\bf{n}}
\end{equation}
Here ${\bf{A}} \in {{\bf{R}}^{M \times N}}$ is an underdetermined
measurement matrix with low coherence of columns, ${\bf{n}} \in
{{\bf{R}}^M}$ represents the acquisition noise vector and ${\bf
x}_0$ a sparse signal. For the inverse problem, both ${\bf x}_0$
and $\bf n$ are unknown, and our goal is to recover ${\bf x}_0$
from $\bf y$. We formulate the observation noise as zero mean
independent Gaussian variable with variance ${\xi ^2}$ i.e.
${\bf{y}}\sim N({{\bf{A}}{{{\bf x}_{0}}}},{\xi ^2}\bf{I})$ and
seek a sparse signal ${\bf x}\buildrel \Delta \over =
({x_1},{x_2}, \ldots {x_N})$ for ${\bf x}_0$.\par In this section,
we utilize a two-stage hierarchical bayesian prior for the signal
model. In the first layer of signal model, we use the Normal
Product (NP) distribution as sparseness prior:
$${\bf{x}} \sim NP({\bf{0}},{\bf{\Sigma }})$$
where ${\bf{0}} \in {{\bf{R}}^N}$ stands for its mean and ${\bf{\Sigma }}$ given as a diagonal matrix ${\text {diag}}({\sigma _1}^2,{\sigma _2}^2, \ldots ,{\sigma _N}^2) \in {{\bf{R}}^{N \times N}}$ represents its variance. For each element ${x_i}$ of ${\bf{x}} $ , its probability density distribution is ${{{K_0}\left( {\frac{{|{x_i}|}}{{{\sigma _i}}}} \right)} \mathord{\left/
 {\vphantom {{{K_0}\left( {\frac{{|{x_i}|}}{{{\sigma _i}}}} \right)} {\pi {\sigma _i}}}} \right.
 \kern-\nulldelimiterspace} {\pi {\sigma _i}}}$, which exhibits a sharp peak at the origin and heavy tails, where ${K_0}( \cdot )$ is the zero order modified Bessel function of the second kind\cite{NP} thus the probability density function (PDF) of $\bf x$ is
 \begin{equation}
   \frac{1}{{{\pi ^N}}}\prod\nolimits_{i = 1}^N {\frac{1}{{{\sigma _i}}}{K_0}(\frac{{\left| {{x_i}} \right|}}{{{\sigma _i}}})}
.\end{equation}
  A NP distributed scaler variate $x_i \in {\bf{R}}$ can be decomposed as the product of two independent normally distributed variables i.e. if $x_i \sim NP(0,\sigma_i^2 )$, then there are $a_i \sim N(0,\kappa_i^2 )$ and $b_i \sim N(0,\gamma_i^2 )$
 satisfying $x_i = a_ib_i$ with moment relationship $\sigma_i  = \gamma_i \kappa_i $\cite{seijas2012approach}. We call this property as the generating rule of Normal Product distribution. Similarly for the vector ${\bf{x}}$,
 we can decompose it into the Hadamard product of two independent virtual normally distributed vector variables ${\bf{a}}$ and ${\bf{b}}$ whose
variance are diagonal matrix ${{\boldsymbol{\kappa }}^2}  \buildrel \Delta \over ={\text{diag}}(\kappa _1^2,\kappa _2^2, \ldots ,\kappa _N^2)$ and ${{\boldsymbol{\gamma }}^2} \buildrel \Delta \over = {\text{diag}}(\gamma _1^2,\gamma _2^2, \ldots ,\gamma _N^2)$ respectively i.e. ${\boldsymbol{x }} = {{\boldsymbol{a }}} \circ {{\boldsymbol{b }}}$ and ${\boldsymbol{\Sigma }} = {{\boldsymbol{\kappa }}^2} \circ {{\boldsymbol{\gamma }}^2}$, where $\circ$ denote the Hadamard product.
Finally, we set $\sigma _i^{ - 2}$ as a realization of Gamma hyperprior $\Gamma (\alpha ,\beta )$ and choose $\alpha  \to 0$,$\beta  \to 0$ to construct a sharper distribution of ${\bf{x}}$ during the variational procedure \cite{wipf2004perspectives} for the second layer of signal.\par
According to the generating rule of Normal Product mentioned above, we can add two parallel nodes ${\bf{a}}$ and ${\bf{b}}$ to construct one latent
 layer. Then the posterior distribution can be expressed as:
\begin{equation*}
{P_{\bf{x}}}({\bf{x}},{\bf{\Sigma }}|{\bf{y}};\alpha ,\beta ) = \int {P({\bf{x}},{\bf{a}},{\bf{b}},{\bf{\Sigma }}|{\bf{y}};\alpha ,\beta )d{\bf{a}}d{\bf{b}}}
\end{equation*}
where
\[\begin{array}{l}
P({\bf{x}},{\bf{a}},{\bf{b}},{\bf{\Sigma }}|{\bf{y}};\alpha ,\beta )\\
 \propto P({\bf{y}}|{\bf{x}})P({\bf{x}}|{\bf{a}},{\bf{b}})P({\bf{a}}|{{\bf{\kappa }}^2})P({\bf{b}}|{{\bf{\gamma }}^2})P({\bf{\Sigma }}|\alpha ,\beta )\\
 = P({\bf{y}} - {\bf{Ax}})\delta ({\bf{x}} - {\bf{a}} \circ {\bf{b}})P({\bf{a}}|{{\bf{\kappa }}^2})P({\bf{b}}|{{\bf{\gamma }}^2})P({\bf{\Sigma }}|\alpha ,\beta )
\end{array}\]
and $\delta(\cdot)$ denotes the Dirac Delta function. In this expression we know that the value of ${\bf{x}}$ must
be equal to ${\bf{a}} \circ {\bf{b}}$ while keeping the value of $P({\bf{x}},{\bf{a}},{\bf{b}},{\bf{\Sigma }}|{\bf{y}};\alpha ,\beta )$ nonzero.
Consider the Bayesian Risk function:
\[\begin{array}{l}
E(L({\bf{\hat x}},{\bf{x}})) = \int {L({\bf{\hat x}},{\bf{x}})} {P_{\bf{x}}}({\bf{x}},{\bf{\Sigma }}|{\bf{y}};\alpha ,\beta )d{\bf{x}}d{\bf{\Sigma }}\\
 \propto \int {L({\bf{\hat x}},{\bf{a}} \circ {\bf{b}})} {P_{{\bf{a}},{\bf{b}}}}({\bf{a}},{\bf{b}},{\bf{\Sigma }}|{\bf{y}};\alpha ,\beta )d{\bf{a}}d{\bf{b}}d{\bf{\Sigma }}
\end{array}\]
where
\[\begin{array}{l}
{P_{{\bf{a}},{\bf{b}}}}({\bf{a}},{\bf{b}},{\bf{\Sigma }}|{\bf{y}};\alpha ,\beta )\\
 = P({\bf{y}} - {\bf{A}}({\bf{a}} \circ {\bf{b}}))P({\bf{a}}|{{\boldsymbol{\kappa }}^2})P({\bf{b}}|{{\boldsymbol{\gamma }}^2})P({\bf{\Sigma }}|\alpha ,\beta )
\end{array}\]
and $L({\bf{\hat x}},{\bf{x}})$ represents the loss function.
Thus, we can replace ${P_{\bf{x}}}({\bf{x}},{\bf{\Sigma }}|{\bf{y}};\alpha ,\beta )$ with ${P_{{\bf{a}},{\bf{b}}}}({\bf{a}},{\bf{b}},{\bf{\Sigma }}|{\bf{y}};\alpha ,\beta )$
 in the inference procedure while maintaining the value of the Bayesian risk function at a consistent level. So the modified Bayesian Network is:

\begin{equation}\label{network observation}
{\bf{y}}\sim N({\bf{A}}({\bf{a}} \circ {\bf{b}}),{\xi ^2}{\bf{I}})
\end{equation}
\begin{equation}\label{network_a}
{\bf{a}} \sim N({\bf{0}},{{\boldsymbol{\kappa }}^2})
\end{equation}
\begin{equation}\label{network_b}
{\bf{b}} \sim N({\bf{0}},{{\boldsymbol{\gamma }}^2})
\end{equation}
\begin{equation}\label{network_gamma}
\sigma _i^{ - 2} = \gamma _i^{ - 2}\kappa _i^{ - 2} \sim \Gamma (\alpha ,\beta )
\end{equation}
as depicted in Fig. \ref{bayesian network graph}.
\begin{figure}[!ht]
\centering
\includegraphics[width=0.5\textwidth]{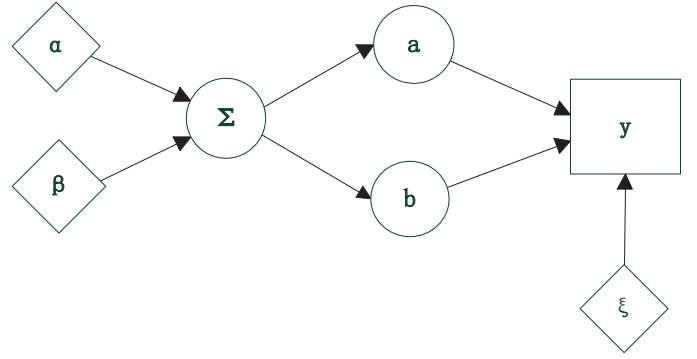}
\caption{The Bayesian Network}
\label{bayesian network graph}
\end{figure}

\section{The Varational Bayesian inference}\label{bayesian inference}
In order to infer the bayesian network, we use the mean-field
variational Bayesian method to analytically approximate the entire
posterior distribution of latent variable. In this theorem, it was
assumed that the approximate variational distribution
$q({\bf{z}}|{\bf{y}})$ where $\bf z$ stands for the latent
variable in the model could be factorized into the product
$\prod\nolimits_{i = 1}^l {{q_i}({{\bf{z}}_i}|{\bf{y}})} $ for the
partition ${{\bf{z}}_1},\ldots, {{\bf{z}}_l}$ of ${\bf{z}}$. It
could be shown that the solution of the variational method for
each factor ${q_i}({{\bf{z}}_i}|{\bf{y}})$ could be written as:
$${q_i}^*({{\bf{z}}_i}|{\bf{y}}) = \frac{{{e^{ < \ln p({\bf{z}}|{\bf{y}})){ > _{ \ne {{\bf{z}}_i}}}}}}}{{\int {{e^{ < \ln p({\bf{z}}|{\bf{y}})){ > _{ \ne {{\bf{z}}_i}}}}}d{{\bf{z}}_i}} }}$$
where $ < \ln p({\bf{z}}|{\bf{y}})){ > _{ \ne {{\bf{z}}_i}}}$ means taking the expectation over the set of ${\bf{z}}$ using the variational distribution except ${{\bf{z}}_i}$\cite{tzikas2008variational} .
    Applying the variational Bayesian method upon the bayesian model mentioned in secetion \ref{bayesian network}, the posterior approximation of ${\bf{\Sigma }},{\bf{a}},{\bf{b}}$ are respectively:
\begin{equation}\label{a}
\ln ({q^*}({\bf{a}})) \propto  < \ln P({\bf{a}},{\bf{b}},{\bf{\Sigma }}|{\bf{y}};{\bf{\alpha }},{\bf{\beta }}){ > _{{q^*}({\bf{b}}),{q^*}({\bf{\Sigma }})}}
\end{equation}
\begin{equation}\label{b}
\ln ({q^*}({\bf{b}})) \propto  < \ln P({\bf{a}},{\bf{b}},{\bf{\Sigma }}|{\bf{y}};{\bf{\alpha }},{\bf{\beta }}){ > _{{q^*}({\bf{a}}),{q^*}({\bf{\Sigma }})}}
\end{equation}
\begin{equation}\label{ga}
  \ln ({q^*}({\bf{\Sigma }})) \propto  < \ln P({\bf{a}},{\bf{b}},{\bf{\Sigma }}|{\bf{y}};{\bf{\alpha }},{\bf{\beta }}){ > _{{q^*}({\bf{b}}),{q^*}({\bf{a}})}}
\end{equation}

\subsection{posterior approximation of ${\bf{a}}$ and ${\bf{b}}$}
Substituting (\ref{network observation}), (\ref{network_a}), (\ref{network_b}) and (\ref{network_gamma}) into (\ref{b}), we can arrive at:
\[{q^*}({\bf{b}}){\sim}\;N({{\bf{\Gamma }}'}^{ - 1}{{\bf{c}}'},{{\bf{\Gamma }}'}^{ - 1})\]
where
$${{\bf{\Gamma }}'} = \frac{1}{{{\xi ^2}}}({{\bf{A}}^T}{\bf{A}}) \circ  < {\bf{a}}{{\bf{a}}^T}{ > _{{q^*}({\bf{a}})}} +  < {{\boldsymbol{\gamma }}^{ - 2}}{ > _{{q^*}({\boldsymbol{\gamma }})}}$$ and
$${{\bf{c}}'} = \frac{1}{{{\xi ^2}}}{\text{diag}}\left( { < {\bf{a}}{ > _{{q^*}({\bf{a}})}}} \right){{\bf{A}}^T}\boldsymbol{y}$$.
From the principle of variational inference, we know that $q\left( {\boldsymbol{b}} \right)$ is an approximation of $P(\boldsymbol{b}|\boldsymbol{y})$ i.e. $P(\boldsymbol{b}|\boldsymbol{y}) \approx q\left( {\bf{b}} \right)$ and in order to get a more concise iteration formula, we choose the relaxation of the secondary moment as:

\[ < {\bf{a}}{{\bf{a}}^T}{ > _{{q^{*}}(a)}} \approx  < {\bf{a}}{ > _{{q^{**}}(a)}} < {{\bf{a}}^T}{ > _{{q^{**}}(a)}}\]
\[ < {{\boldsymbol{\gamma }}^{ - 2}}{ > _{{q^*}({\boldsymbol{\gamma }})}} \approx  < {\boldsymbol{\gamma }} > _{{q^{**}}({\boldsymbol{\gamma }})}^{ - 2}\]

This relaxation can be interpreted as ignoring the posteriori variance during the updating, which is inspired by the fact that after the learning procedure
 the posteriori variance always approaches zero to ensure the posteriori mean's concentration on the estimated value.

Then the corrected posterior approximation is ${q^{**}}({\bf{b}})\sim N({{\boldsymbol{\Gamma }}^{ - 1}}{\bf{c}},{{\boldsymbol{\Gamma }}^{ - 1}})$ where
\[\begin{array}{*{20}{l}}
{{{\boldsymbol{\Gamma }}^{ - 1}} = \{ \frac{1}{{{\xi ^2}}}{{\left( {{\bf{A}}{\rm{diag}}\left( { < {\bf{a}}{ > _{{q^{**}}({\bf{a}})}}} \right)} \right)}^T}{\bf{A}}{\rm{diag}}\left( { < {\bf{a}}{ > _{{q^{**}}({\bf{a}})}}} \right)}\\
{{\rm{ + }} < {\boldsymbol{\gamma }}{ > _{{q^{**}}({\boldsymbol{\gamma }})}}^{ - 2}{\} ^{ - 1}}}
\end{array}\]
and
\[{{\bf{c}} = \frac{1}{{{\xi ^2}}}{\rm{diag}}\left( { < {\bf{a}}{ > _{{q^{**}}({\bf{a}})}}} \right){{\bf{A}}^T}{\bf{y}}}\]
In the noise free case, using Woodbury identity we have:
\begin{equation}\label{b_mean}
\begin{array}{*{20}{l}}
{\mathop {\lim }\limits_{\xi  \to 0} {{\boldsymbol{\Gamma }}^{ - 1}}{\bf{c}}}\\
{ =  < {\boldsymbol{\gamma }}{ > _{{q^{**}}({\boldsymbol{\gamma }})}}{{({\bf{A}}{\rm{diag}}( < {\bf{a}}{ > _{{q^{**}}({\bf{a}})}}) < {\boldsymbol{\gamma }}{ > _{{q^{**}}({\boldsymbol{\gamma }})}})}^ + }{\bf{y}}}
\end{array}
\end{equation}
and
\begin{equation}\label{b_variance}
\begin{array}{l}
{\lim _{\xi  \to 0}}{{\bf{\Gamma }}^{ - 1}}\\
 = (\bf{I} -  < {\boldsymbol{\gamma }}{ > _{{q^{**}}({\boldsymbol{\gamma }})}}{\left( {{\bf{A}}{\text{diag}}\left( { < {\bf{b}}{ > _{{q^{**}}({\bf{b}})}}} \right) < {\boldsymbol{\gamma }}{ > _{{q^{**}}({\boldsymbol{\gamma }})}}} \right)^ + })\\
 \times {\bf{A}}{\text{diag}}\left( { < {\bf{a}}{ > _{{q^{**}}({\bf{a}})}}} \right) < {\boldsymbol{\gamma }} > _{{q^{**}}({\boldsymbol{\gamma }})}^2
\end{array}
\end{equation}
Similarly, for ${\bf{a}}$, substituting (\ref{network observation}), (\ref{network_a}), (\ref{network_b}) and (\ref{network_gamma}) into (\ref{a}), we have:
\[{q^*}({\bf{a}}){\sim}\;N({{\bf{{ H}}}'}^{ - 1}{{\bf{f}}'},{{\bf{{ H}}}'}^{ - 1})\]
where
\[{{\bf{{ H}}}'} = \frac{1}{{{\xi ^2}}}({{\bf{A}}^T}{\bf{A}}) \circ  < {\bf{b}}{{\bf{b}}^T}{ > _{{q^*}({\bf{b}})}} +  < {{\boldsymbol{\kappa }}^{ - 2}}{ > _{{q^*}({\boldsymbol{\kappa }})}}\]
and
$${{\bf{f}}'} = \frac{1}{{{\xi ^2}}}{\text{diag}}\left( { < {\bf{b}}{ > _{{q^*}({\bf{b}})}}} \right){{\bf{A}}^T}\bf{y}$$
So the corrected approximation is:
$${q^{**}}({\bf{a}})\sim N({{\bf{{ H}}}^{ - 1}}{\bf{f}},{{\bf{ H}}^{ - 1}})$$
Then again, in the noiseless case, we have
\begin{equation}\label{a_mean}
\begin{array}{l}
{\lim _{\xi  \to 0}}{{\bf{{ H}}}^{ - 1}}{\bf{f}}\\
 =  < {\boldsymbol{\kappa }}{ > _{{q^{**}}({\boldsymbol{\kappa }})}}{({\bf{A}}{\text{diag}}( < {\bf{b}}{ > _{{q^{**}}({\bf{b}})}}) < {\boldsymbol{\kappa }}{ > _{{q^{**}}({\boldsymbol{\kappa }})}})^ + }\bf{y}
\end{array}
\end{equation}
and
\begin{equation}\label{a_variance}
\begin{array}{l}
{\lim _{\xi  \to 0}}{{\bf{H}}^{ - 1}}\\
 = (\bf{I} -  < {\boldsymbol{\kappa }}{ > _{{q^{**}}({\boldsymbol{\kappa }})}}{\left( {{\bf{A}}{\text{diag}}\left( { < {\bf{b}}{ > _{{q^{**}}({\bf{b}})}}} \right) < {\boldsymbol{\kappa }}{ > _{{q^{**}}({\boldsymbol{\kappa }})}}} \right)^ + })\\
 \times {\bf{A}}{\text{diag}}\left( { < {\bf{b}}{ > _{{q^{**}}({\bf{b}})}}} \right) < {\boldsymbol{\kappa }} > _{{q^{**}}({\boldsymbol{\kappa }})}^2
\end{array}
\end{equation}

\subsection{posterior approximation of ${\bf{\Sigma}}$ }

As discussed in section \ref{bayesian network}, we have:
\[\begin{array}{c}
\ln ({q^*}({\bf{\Sigma }})) = \ln ({q^*}({{\boldsymbol{\kappa }}^2} \circ {{\boldsymbol{\gamma }}^2})) = \ln ({q^*}({{\boldsymbol{\kappa }}^2})) + \ln ({q^*}({{\boldsymbol{\gamma }}^2}))\\
 \propto  < \ln P({\bf{a}},{\bf{b}},{{\boldsymbol{\kappa }}^2} \circ {{\boldsymbol{\gamma }}^2}|{\bf{y}};{\boldsymbol{\alpha }},{\boldsymbol{\beta }}){ > _{{q^{**}}({\bf{b}}),{q^{**}}({\bf{a}})}}
\end{array}\]
Substituting (\ref{network_a}), (\ref{network_b}), (\ref{network_gamma}) into (\ref{ga}) and using the separability of $ \ln ({q^*}({{\boldsymbol{\kappa }}^2}))$, it can be shown that:
$$\begin{array}{c}
lnq\left( {\kappa _i^2} \right) \propto  - \frac{1}{2}\left( { < a_i^2{ > _{{q^{**}}({\bf{a}})}} + 2\beta  < {\gamma _i}^{ - 2}{ > _{{q^*}(\gamma _i^2)}}} \right){\kappa _i}^{ - 2}\\
 + (\alpha  + \frac{1}{2} - 1)\ln \left( {{\kappa _i}^{ - 2}} \right)
\end{array}$$
which means
$${\kappa _i}^{ - 2} \sim \Gamma (\frac{1}{2} + \alpha ,\frac{1}{2}( < a_i^2{ > _{{q^{**}}({\bf{a}})}} + 2\beta  < {\gamma _i}^{ - 2}{ > _{{q^*}({\boldsymbol{\gamma }}^2)}}))$$
and
\begin{equation}\label{meankappa}
< {\kappa _i}^{ - 2}{ > _{{q^*}({\boldsymbol{\kappa} ^2})}} = \frac{{\frac{1}{2} + \alpha }}{{\frac{1}{2}( < a_i^2{ > _{{q^{**}}({\bf{a}})}} + 2\beta  < {\gamma _i}^{ - 2}{ > _{{q^*}({\boldsymbol{\gamma }}^2)}})}}
\end{equation}

Similarly, we have:
$${\gamma _i}^{ - 2} \sim \Gamma (\frac{1}{2} + \alpha ,\frac{1}{2}( < b_i^2{ > _{{q^{**}}({\bf{b}})}} + 2\beta  < {\kappa _i}^{ - 2}{ > _{{q^*}({\boldsymbol{\kappa }}^2)}}))$$
and
\begin{equation}\label{meangamma}
 < {\gamma _i}^{ - 2}{ > _{{q^*}({\boldsymbol{\gamma}}^2 )}} = \frac{{\frac{1}{2} + \alpha }}{{\frac{1}{2}( < b_i^2{ > _{{q^{**}}({\bf{b}})}} + 2\beta  < {\kappa _i}^{ - 2}{ > _{{q^*}({\boldsymbol{\kappa }}^2)}})}}
\end{equation}

\subsection{The Proposed Algorithm }
As a result, we summarize the procedure above as two algorithms
named ``NP-0'' and ``NP-1'' which represent the inference results
for the one layer signal model and two layer signal model
respectively. The difference between Np-0 and NP-1 is whether the
learning process updates the precision of Normal Product.

\begin{figure}[h]
\centering
\footnotesize
\fbox{\parbox{\hsize}{
Algorithm 1: NP-0 (One Layer Normal Product)
}}
\renewcommand{\algorithmicrequire}{\textbf{Input:}}
\renewcommand\algorithmicensure {\textbf{Output:} }
\fbox{\parbox{\hsize}{
\begin{algorithmic}[1]
\REQUIRE $\bf{A}$,$\bf{y}$
\ENSURE ${\bf{x}}^{(t)}$

\STATE $ < {\bf{a}}{ > ^{(0)}} \leftarrow \mathbf{1}$, $ < {\bf{b}}{ > ^{(0)}} \leftarrow \mathbf{1}$,${ < {\boldsymbol{\gamma }} > } \leftarrow \mathbf{1}$,${ < {\boldsymbol{\kappa }} > }\leftarrow \mathbf{1}$, $\varepsilon $, $t_{max}$,
\WHILE{$\left\| {{x^{(t)}} - {x^{(t - 1)}}} \right\|/\left\| {{x^{(t - 1)}}} \right\| > \varepsilon $ and $t<t_{max}$}
\STATE Update $ < {\bf{a}}{ > ^{(t)}}$ using (\ref{a_mean})
\STATE Update $ < {\bf{b}}{ > ^{(t)}}$ using (\ref{b_mean})
\STATE ${{\bf{x}}^{(t)}} =  < {\bf{a}}{ > ^{(t)}} \circ  < {\bf{b}}{ > ^{(t)}}$
\STATE $ t=t+1$
\ENDWHILE

\end{algorithmic}
}}
\end{figure}
The algorithm NP-0 is similar to the FOCUSS algorithm\cite{gorodnitsky1997sparse}. The difference between them is that NP-0 uses $<{\bf a}>^{(t)}$ and $<{\bf b}>^{(t)}$, the decomposed component of ${\bf x}^{(t)}$ in an interleaved way to update ${\bf x}^{(t+1)}$ while FOCUSS uses the whole ${\bf x}^{(t)}$ to directly update ${\bf x}^{(t+1)}$. Furthermore, we set the initial value of the algorithm as a constant to avoid the local minimum results being returned.

\begin{figure}[h]
\centering
\footnotesize
\fbox{\parbox{\hsize}{
Algorithm 2: NP-1 (Two Layers Normal Product)
}}
\renewcommand{\algorithmicrequire}{\textbf{Input:}}
\renewcommand\algorithmicensure {\textbf{Output:} }
\fbox{\parbox{\hsize}{
\begin{algorithmic}[1]
\REQUIRE $\bf{A}$,$\bf{y}$
\ENSURE ${\bf{x}}^{(t)}$

\STATE $ < {\bf{a}}{ > ^{(0)}} \leftarrow \mathbf{1}$, $ < {\bf{b}}{ > ^{(0)}} \leftarrow \mathbf{1}$,$\alpha\leftarrow 0$, $\beta\leftarrow 0$, $\varepsilon $, $t_{max}$

\WHILE{$\left\| {{{\bf x}^{(t)}} - {{\bf x}^{(t - 1)}}} \right\|/\left\| {{{\bf x}^{(t - 1)}}} \right\| > \varepsilon $ and $t<t_{max}$}
\STATE Update $ < {\bf{a}}{ > ^{(t)}}$ using (\ref{a_mean})
\STATE Update $ Var({\bf{a}}) ^{(t)}$ using (\ref{a_variance})
\STATE Update $ < {\bf{b}}{ > ^{(t)}}$ using (\ref{b_mean})
\STATE Update $ Var({\bf{b}}) ^{(t)}$ using (\ref{b_variance})
\STATE ${{\bf{x}}^{(t)}} =  < {\bf{a}}{ > ^{(t)}} \circ  < {\bf{b}}{ > ^{(t)}}$
\STATE Update $ < {\kappa _i}^{ - 2} > _{_{{q^*}({{\bf{\kappa }}^2})}}^{(t)}$ using (\ref{meankappa})
\STATE Update $ < {\gamma _i}^{ - 2} > _{{q^*}({{\bf{\gamma }}^2})}^{(t)}$ using (\ref{meangamma})
\STATE $ t=t+1$
\ENDWHILE
\end{algorithmic}
}}
\end{figure}

The algorithm NP-1 looks like coupling two sparse Bayesian learning (SBL) \cite{tipping2001sparse,wipf2004sparse} procedures together and we will experientially
prove that the MSE via NP-1 descends faster than SBL in section \ref{discussion}.

\section{Numerical Results}\label{discussion}
In this section, we compare our proposed algorithms among SBL, iterative reweighted least squares (IRLS)\cite{chartrand2008iteratively} and basis pursuit (BP)\cite{chen1998atomic}.
In the following results, we set the dimension of original signal to be 100 and in each experiment every entries of the sensing matrix $\bf A$ is uniformly distributed as standard normal random variable. The stop criterion parameters $\varepsilon $ and $t_{max}$ in both algorithm NP-0 and NP-1 are set as $10^{-3}$ and $300$ respectively.\par
\begin{figure}[th!]
\centering
\subfloat[K=3]{
\includegraphics[width=4.3cm,height=4.3cm]{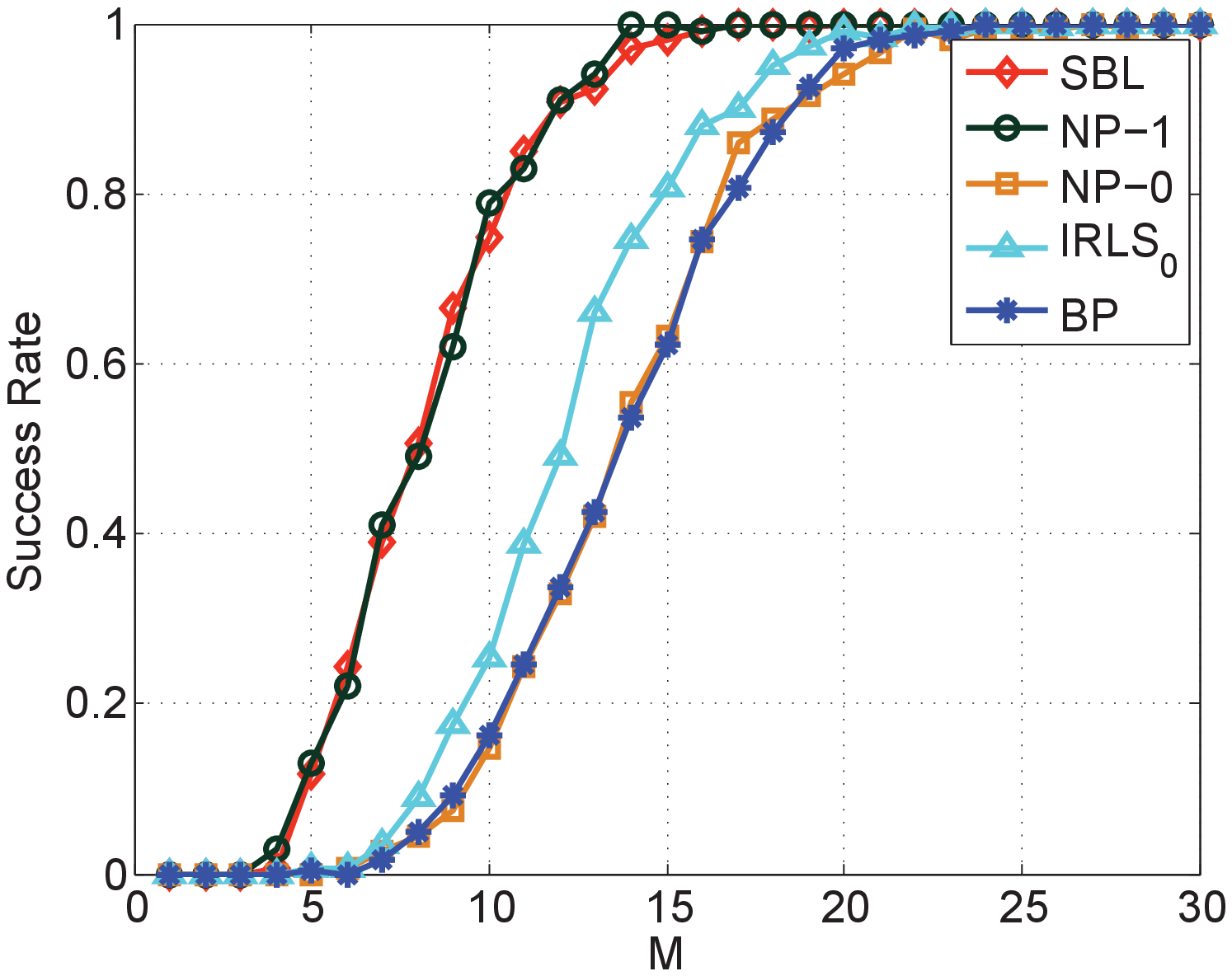}%
\label{succK=3}}
\hfil
\subfloat[M=30]{
\includegraphics[width=4.3cm,height=4.4cm]{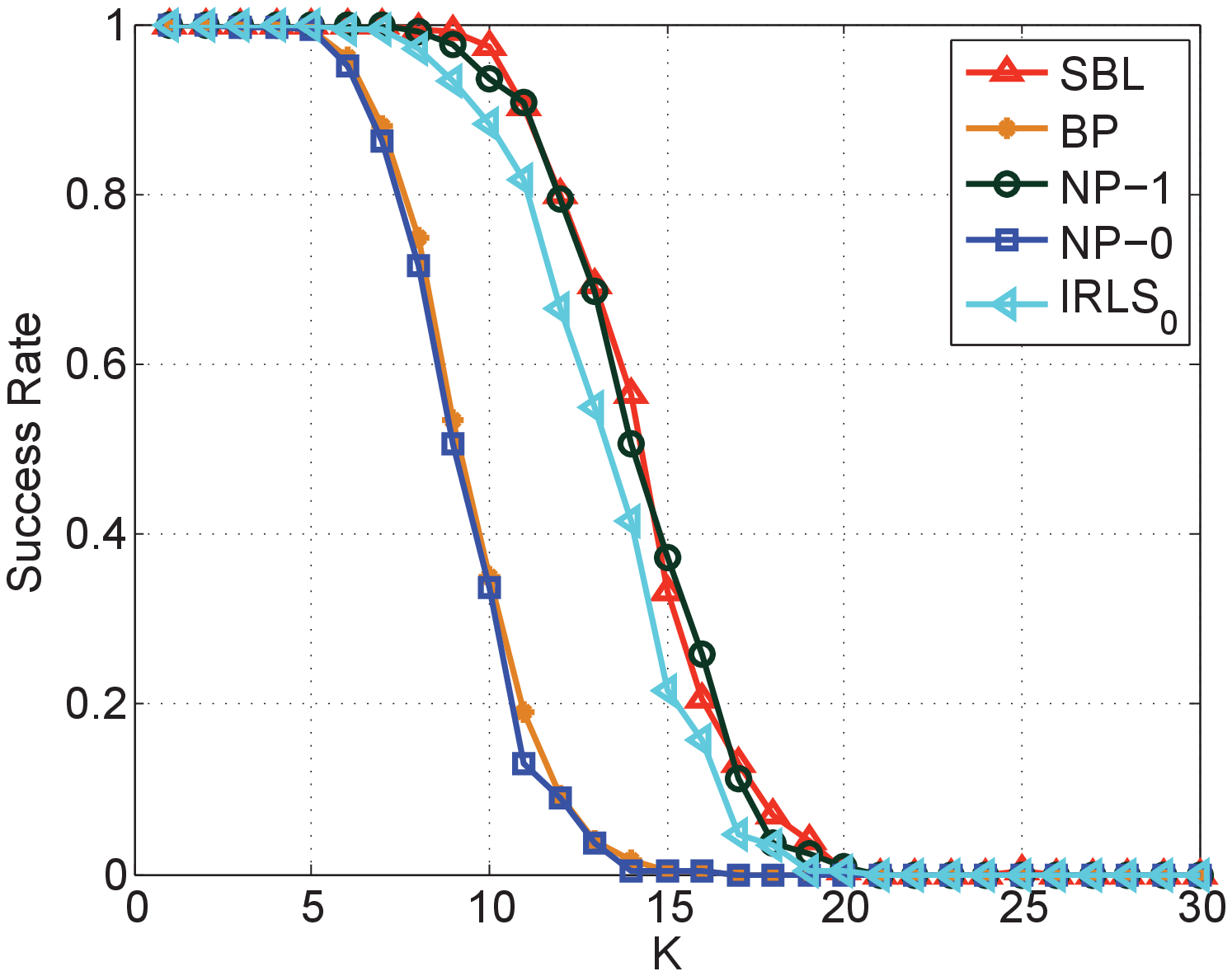}%
\label{succM=30}}
\caption{Simulation results of Success Rate}
\label{successrate}
\end{figure}
\begin{figure}[th!]
\centering
\subfloat[K=3,M=30]{
\includegraphics[width=4.3cm,height=4.3cm]{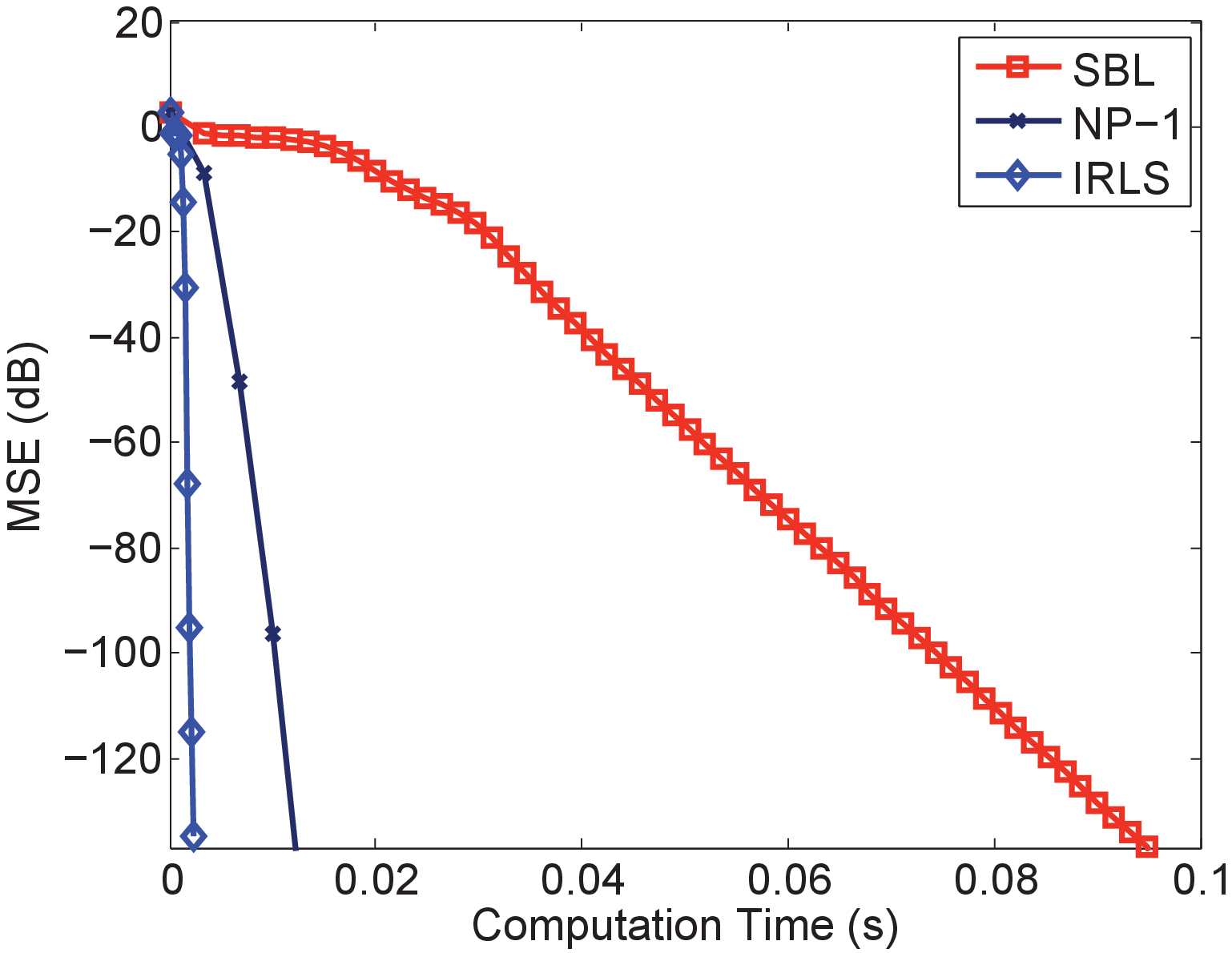}%
\label{msecomputationK=3,M=30}}
\hfil
\subfloat[K=3,M=30]{
\includegraphics[width=4.3cm,height=4.3cm]{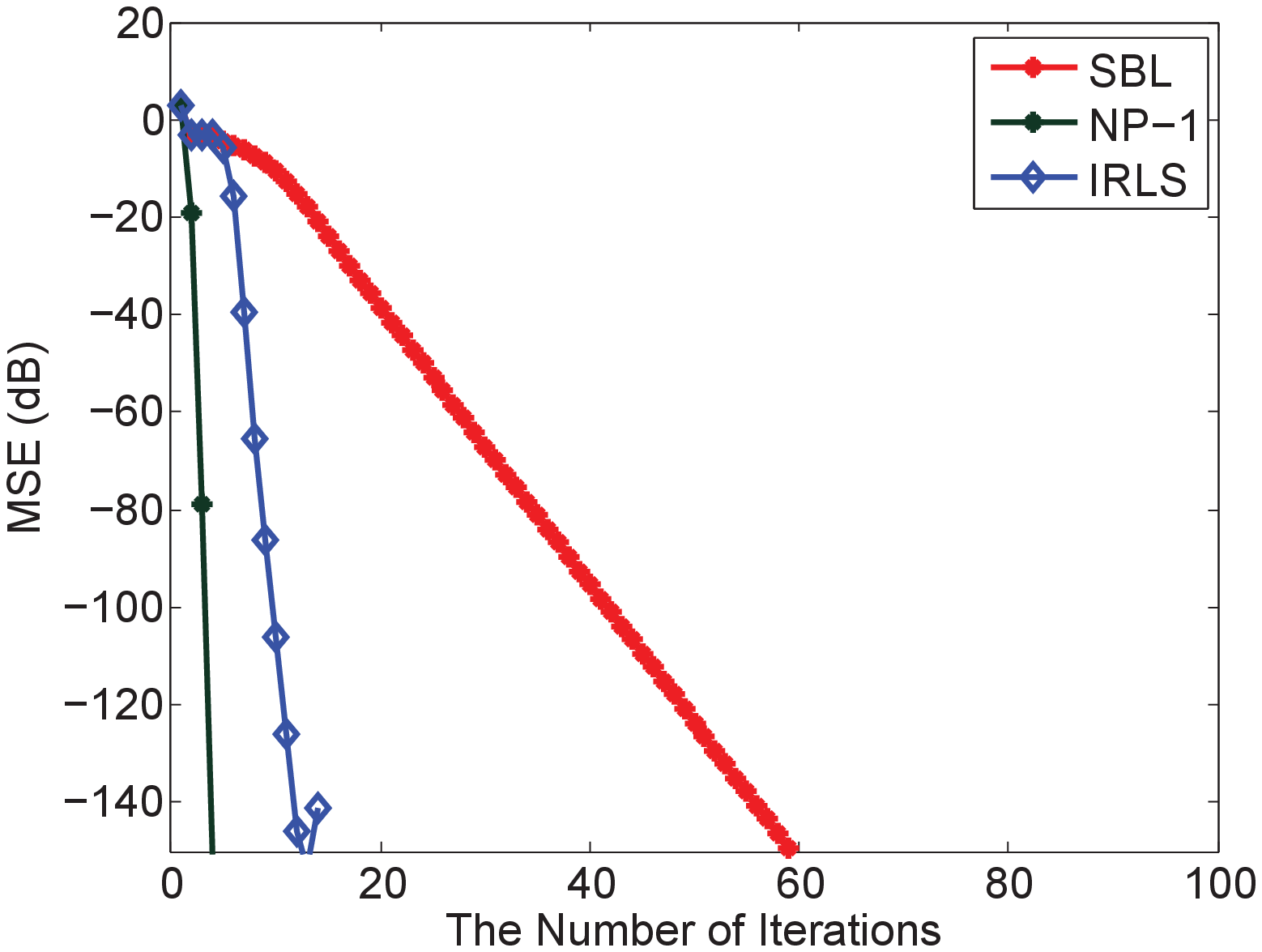}%
\label{mseiterationk=3,M=30}}
\caption{Simulation results of Computational Cost}
\label{computation}
\end{figure}

The success rate in the Fig. \ref{successrate} is defined by the average probability of successful restoration, which is counted as the ratio between the number of successful trials and the total number of the independent experiments. It collects the result of 300 Monte Carlo trials and
we assume successful trial if ${{{{\left\| {{\bf{\hat x}} - {\bf{x}}} \right\|}_2}} \mathord{\left/
 {\vphantom {{{{\left\| {{\bf{\hat x}} - {\bf{x}}} \right\|}_2}} {{{\left\| {\bf{x}} \right\|}_2}}}} \right.
 \kern-\nulldelimiterspace} {{{\left\| {\bf{x}} \right\|}_2}}} < {10^{ - 3}}$, where ${{\bf{\hat x}}}$ is the estimate of the original signal $\bf{x}$.
The mean squared error (MSE) in Fig. \ref{computation} which measures the average of the squares of the difference between the estimator and estimated signal is calculated as follow:
\[MSE(dB) = 20\log ({{{{\left\| {{\bf{\hat x}} - {\bf{x}}} \right\|}_2}} \mathord{\left/
 {\vphantom {{{{\left\| {{\bf{\hat x}} - {\bf{x}}} \right\|}_2}} {{{\left\| {\bf{x}} \right\|}_2}}}} \right.
 \kern-\nulldelimiterspace} {{{\left\| {\bf{x}} \right\|}_2}}})\]\par
  Fig. \ref{successrate} demonstrats the superior performance of the NP-1 algorithm in a few measurements. It shows the success rate of respective algorithms versus the number of measurement M and sparse level K of the signal. We can see that the reconstruction performance of the one layer normal product algorithm(NP-0) is almost the same as BP, which is the MAP estimation of unknowns in the one layer Laplace prior framework. The comparisons also show that the two layers normal product algorithm(NP-1) requires as few number of CS measurements as SBL while inheriting the similar reconstruction precision.\par

In Fig. \ref{computation}, the comparison among the computational cost of the three algorithms on the reconstruction performance are performed on a personal PC with dual-core 3.1 GHz CPU and 4GB RAM. It is interesting to note that the MSE of NP-1 descends faster than SBL in Fig. 4(a) while it has the same sparsity-undersampling tradeoff performance as SBL. It should also be noticed that we haven't use any basis pruning technology as in \cite{tipping2003fast} to reduce the computational complexity. To understand the reason of low computational time from Fig. 4(b), we observe that the NP-1 requires only a few numbers of iterations to arrive at stop condition. In summary, these experimental results confirm that the NP-1 is a fast Sparse Bayesian algorithm.

\section{Conculsion}

In this letter, we formulated a Normal Product prior based Bayesian framework to solve the compressed sensing problem in noise free case. Using this framework, we derived two algorithms named NP-0 and NP-1 and compared them with different algorithms. We have shown that our algorithm NP-1 has the similar reconstruction performance with SBL while the interleaved updating procedure provides improved performance in computational times.






\bibliographystyle{IEEEtran}
\bibliography{./ref/NormalProductReference}


\end{document}